\documentclass[conference,a4paper]{IEEEtran}
\usepackage{cite}
\usepackage{amsmath,amssymb,amsfonts}
\usepackage{algorithmic}
\usepackage{graphicx}
\graphicspath{{pics/}}
\usepackage{textcomp}
\usepackage[pdftex,dvipsnames]{xcolor}
\usepackage{etoolbox}
\usepackage[T1]{fontenc}
\usepackage[latin1]{inputenc}
\usepackage[english]{babel}
\usepackage{float}

\usepackage{hyperref}
\addto\extrasenglish{
}

\usepackage{xargs} 

\begin{document}
%
\makeatletter
\g@addto@macro\@maketitle{
  \begin{figure}[H]
  \setcounter{figure}{0}
  \setlength{\linewidth}{\textwidth}
  \setlength{\hsize}{\textwidth}
  \centering
  \includegraphics[width=18cm]{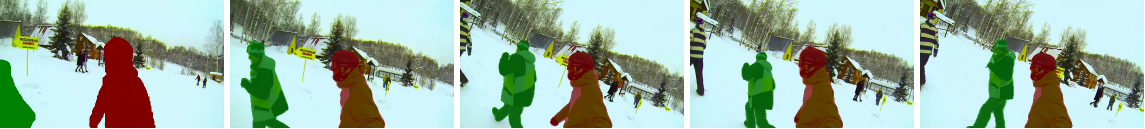}
  \caption{In one-shot video object segmentation, the binary masks of the first appearance of the target objects are provided. 
    The task is to segment these objects throughout the whole sequence. Here the first image on the left shows the initial frame and masks.
    The other segmentation masks are the outputs of our method for the next frames in the sequence. Every two frames are shown here for more coverage of the sequence.}
  \label{fig:intro}
  \end{figure}
}
\makeatother
\title{Revisiting Sequence-to-Sequence Video Object Segmentation with Multi-Task Loss and Skip-Memory}
%


\author{
\IEEEauthorblockN{
Fatemeh Azimi\IEEEauthorrefmark{1}\textsuperscript{,1,2},
Benjamin Bischke\IEEEauthorrefmark{1}\textsuperscript{,1,2},
Sebastian Palacio\textsuperscript{1,2},
Federico Raue\textsuperscript{2},
J{\"o}rn Hees\textsuperscript{2},
Andreas Dengel\textsuperscript{1,2}
}
\IEEEauthorblockA{
\textsuperscript{1}DFKI GmbH\\
\textsuperscript{2}TU Kaiserslautern \\
Kaiserslautern, Germany \\
\url{firstname.lastname@dfki.de}\\
\IEEEauthorrefmark{1}Equal Contribution
}
}

\maketitle

\begin{abstract}
Video Object Segmentation (VOS) is an active research area of the visual domain.
One of its fundamental sub-tasks is semi-supervised / one-shot learning: given only the segmentation mask for the first frame, the task is to provide pixel-accurate masks for the object over the rest of the sequence.
Despite much progress in the last years, we noticed that many of the existing approaches lose objects in longer sequences, especially when the object is small or briefly occluded.
In this work, we build upon a sequence-to-sequence approach that employs an encoder-decoder architecture together with a memory module for exploiting the sequential data.
We further improve this approach by proposing a model that manipulates multi-scale spatio-temporal information using memory-equipped skip connections.
Furthermore, we incorporate an auxiliary task based on distance classification which greatly enhances the quality of edges in segmentation masks.
We compare our approach to the state of the art and show considerable improvement in the contour accuracy metric and the overall segmentation accuracy.
\end{abstract}
\IEEEpeerreviewmaketitle

%
%
%
%
\section{Introduction}
One-shot Video Object Segmentation (VOS) is the task of segmenting an object of interest throughout a video sequence with many applications in areas such as autonomous systems and robotics.
In this task, the first mask of the object appearance is provided and the model is supposed to segment that specific object during the rest of the sequence.
VOS is a fundamental task in Computer Vision dealing with various challenges such as handling occlusion, tracking objects with different sizes and speed, and drastic motion either from the camera or the object \cite{yao2019video}.
Within the last few years, video object segmentation has received a lot of attention from the community \cite{caelles2017one,perazzi2017learning,voigtlaender2017online,tokmakov2017learning}.
Although VOS has a long history \cite{chang2013video,grundmann2010efficient,marki2016bilateral}, only recently it has resurfaced due to the release of large-scale and specialized datasets \cite{pont20172017,perazzi2016benchmark}.

To solve VOS, a wide variety of approaches have been proposed in the literature ranging from training with static images without using temporal information \cite{caelles2017one} to using optical flow for utilizing the motion information and achieving better temporal consistency \cite{tokmakov2017learning}.
However, the methods relying solely on static images lack temporal consistency and using optical flow is computationally expensive and imposes additional constraints.

With the release of YoutubeVOS \cite{xu2018youtub}, the largest video object segmentation dataset to date, the authors demonstrated that having enough labeled data makes it possible to train a sequence-to-sequence (S2S) model for video object segmentation.
In S2S, an encoder-decoder architecture is used similar to \cite{badrinarayanan2017segnet}.
Furthermore, a recurrent neural network (RNN) is employed after the encoder (referred to as bottleneck) to track the object of interest in a temporally coherent manner.

\begin{figure}
    \centering
    \includegraphics[width=3in]{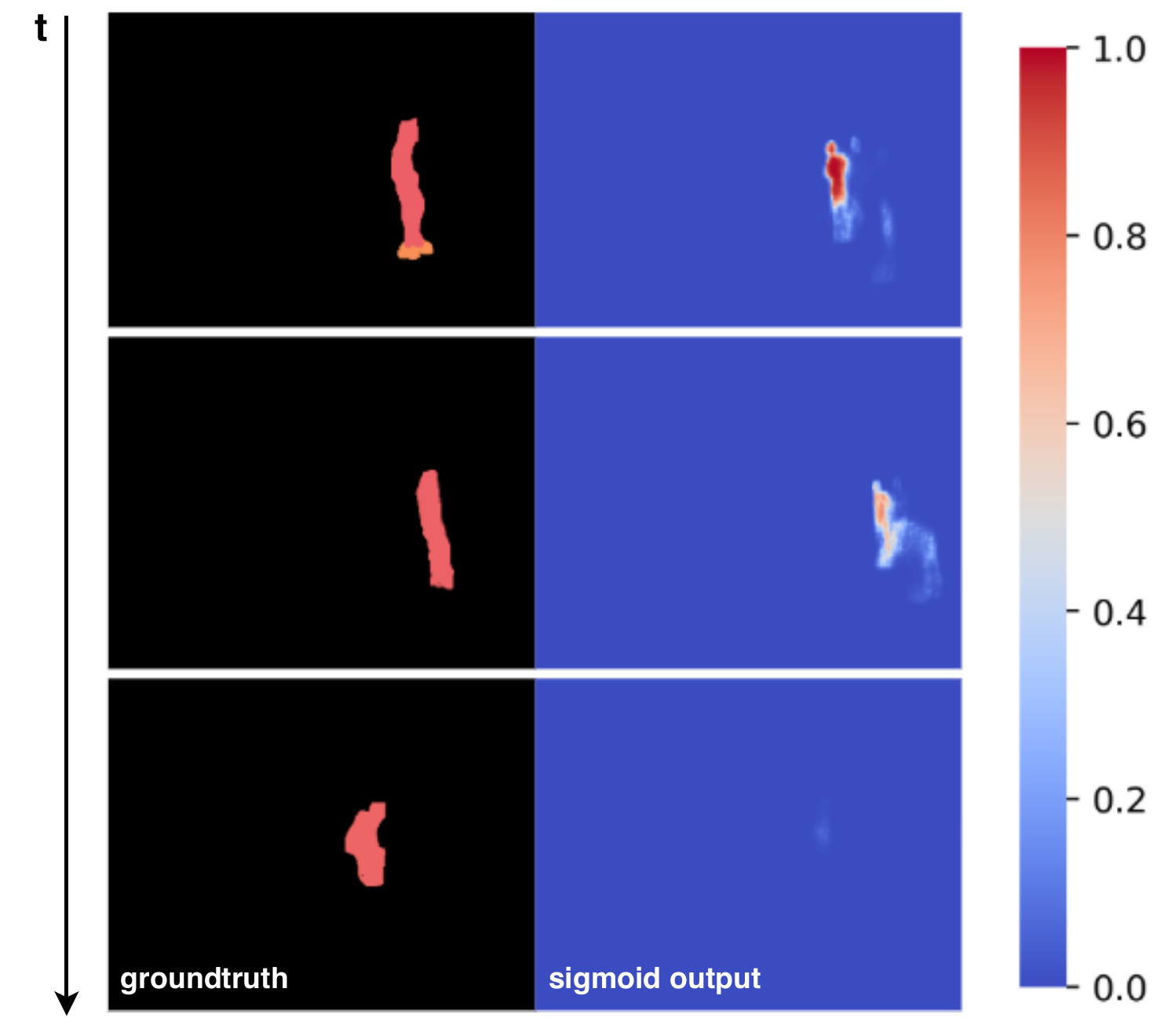}
    \caption{In this image, the ground-truth mask is shown on the left and the output of
    the decoder of the S2S architecture, on the right. 
    The output of the $sigmoid$ function (last layer in the decoder) acts like a probability distribution over the binary classification, measuring the model confidence. The output of around $0.5$ (white color coding) implies a low confidence in the prediction while values close to $0$ or $1$ (blue and red colors) show confident outputs w.r.t. to background and foreground classes).
    Our observation is that the model is not often confident when predicting masks for small objects.
    This uncertainty propagates to the next predictions causing the model to lose the target object within a few time steps.
    We argue that part of this issue is because the RNN located in the bottleneck of the encoder-decoder architecture, does not receive enough information from the small objects. 
    }
    \label{fig:uncertainty}
\end{figure}

In this work, we build on top of the S2S architecture due to its simplicity and elegant design that reaches impressive results compared to the state of the art while remaining efficient \cite{xu2018youtub}.
%
%
In the YoutubeVOS dataset, there are sequences with up to five objects with various sizes to be segmented.
By having a close look at the S2S behavior, we noticed that it often loses track of smaller objects.
The problem in the failure cases is that early in the sequence, the network prediction 
of the segmentation masks has a lower confidence (the output of the $sigmoid$ layer is around $0.5$).
This uncertainty increases and propagates rapidly to the next frames resulting in the model losing the object as shown in \autoref{fig:uncertainty}.
Therefore, the segmentation score of that object would be zero for the rest of the sequence which has a strong negative impact on the overall performance.
We argue that this is partly due to lack of information in the bottleneck, especially for small objects.

To improve the capacity of the model for segmenting smaller objects, we propose utilizing spatio-temporal information at multiple scales.
To this end, we propose using additional skip connections incorporating a memory module (henceforth referred to as skip-memory).
Our intuition is based on the role of ConvLSTM in the architecture that is remembering the area of interest in the image.
Using skip-memory allows the model to track the target object at multiple scales.
This way, even if the object is lost at the bottleneck, there is still a chance to track it by using information from lower scales.

Our next contribution is the introduction of an auxiliary task for improving the performance of video object segmentation.
The effectiveness of multi-task learning has been shown in different scenarios \cite{ruder2017overview},
but it has received less attention in video object segmentation.
We borrow ideas from Bischke et al. \cite{bischke2019multi} for satellite image segmentation and adapt it for the task at hand. 
The auxiliary task defined here is distance classification.
For this purpose, border classes regarding the distance to the edge of the target object is assigned to each pixel in the ground-truth mask.

\begin{figure*}[t!]
    \centering
    \includegraphics[width=15cm]{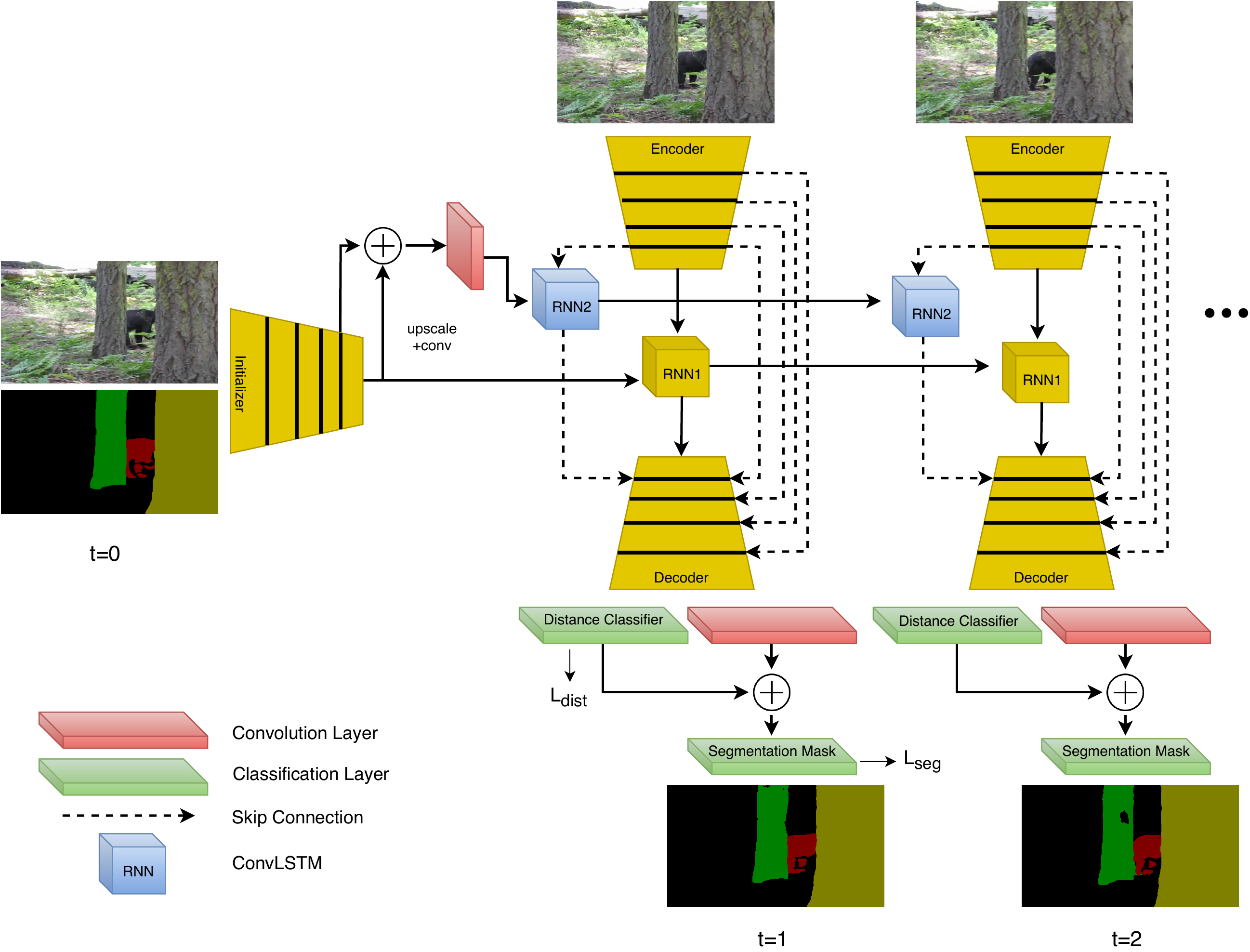}
    \caption{The overall architecture of our approach. We utilize information at different scales of the video by using skip-memory (RNN2). 
    Experiments with more than one skip-memory connection are possible (only one is shown here for simplicity).
    We use an additional distance-based loss to improve the contour quality of the segmentation masks. For this purpose, a distance class is assigned to each pixel in the mask, based on its distance to the object boundary. We use a $softmax$ at the distance classification branch and a $sigmoid$ at the segmentation branch to compute the $L_{dist}$ and $L_{seg}$, respectively.
    Yellow blocks show the architecture of the original S2S model and all other blocks depict our extension to this model.}
    \label{fig:architecture}
\end{figure*}

We adapt the decoder network with an extra branch for the additional distance classification mask and use its output as an additional training signal for predicting the segmentation mask.
The distance classification objective provides more specific information about the precise location of each pixel, resulting in significant improvement of the $F$ score (measuring the quality of the segmentation boundaries).
The overall architecture is shown in \autoref{fig:architecture}.
In the rest of the paper we refer to our method as S2S++.
%
%
%
%
%
\section{Related Work}
\label{seq:related}
One-shot video object segmentation can be seen as pixel-wise tracking of target objects throughout a video sequence, where the first segmentation mask is provided as shown in \autoref{fig:intro}.
This field has a long history in the literature \cite{brox2010object},
however, with the rise of deep learning, classical methods based on energy minimization, using super-voxels and graph-based methods
\cite{papazoglou2013fast,jain2014supervoxel,shankar2015video,faktor2014video} were replaced with deep learning based approaches.
In the following we provide a brief overview of the state of the art approaches in this domain.

Having the first segmentation mask at hand, two training mechanisms exist in the literature:
\emph{offline training} which is the standard training process and \emph{online training} which is performed at the test time.
In online training, heavy augmentation is applied to the first frame of the sequence in order to generate more data and the network is further trained on the specific sequence
\cite{perazzi2017learning,caelles2017one,voigtlaender2017online,Man+18b}.
Using online training as an additional training step leads to better results, however, it makes the inference phase quite slow and computationally more demanding.

Regarding offline training, various approaches have been suggested in the literature.
Some approaches are based on using static images and extending image segmentation for video \cite{caelles2017one,perazzi2017learning}. 
In \cite{caelles2017one}, authors use a VGG architecture \cite{simonyan2014very} pre-trained on ImageNet \cite{krizhevsky2012imagenet} and adapt it for video object segmentation. 
Further offline and online training accompanied by contour snapping, allow the model to keep the object of interest and discard the rest of the image (classified as background). 
\cite{perazzi2017learning} treats the task as guided instance segmentation.
In this case, the previous predicted mask (first mask in the beginning followed by using predicted masks at next time steps) is used as an additional input, serving as the guidance signal. 
Moreover, the authors experiment with different types of signals such as bounding boxes and optical flow demonstrating that even a weak guidance signal such as bounding box can be effective.
In OSMN \cite{yang2018efficient}, the authors propose using two visual and spatial modulator networks to adapt the base network for segmenting only the object of interest.
The main problem with these methods is that they do not utilize sequential data and therefore suffer from a lack of temporal consistency.

Another approach taken in the literature is using object proposals based on RCNN-based techniques \cite{he2017mask}.
In \cite{luiten2018premvos} the task is divided into two steps: First, generating the object proposals and second, selecting and fusing the promising mask proposals trying to enforce the temporal consistency by utilizing optical flow.
In \cite{li2017video} the authors incorporate a re-identification module base on patch-matching to recover from failure cases where the object is lost during segmentation (e.g., as a result of accumulated error and drift in long sequences).
These methods achieve a good performance, with the downside of being quite complex and slow.

Before the introduction of a comprehensive dataset for VOS, it was customary to pre-train the model parameters on image segmentation datasets such as PASCAL VOC \cite{everingham2010pascal} and then fine-tune them on video datasets \cite{voigtlaender2017online,wug2018fast}.
Khoreva et al. suggest an advanced data augmentation method for video segmentation including non-rigid motion, to address the lack of labeled data in this domain \cite{khoreva2019lucid}.
However, with the release of YoutubeVOS dataset \cite{xu2018youtub}, the authors show that it is possible to train an end-to-end, sequence-to-sequence model for video object segmentation when having enough labeled data.
They deploy a ConvLSTM module \cite{xingjian2015convolutional} to process the sequential data and to maintain temporal consistency.

In \cite{tokmakov2017learning}, the authors propose a two-stream architecture composed of an appearance network and a motion network. 
The result of these two branches are merged and fed to a ConvGRU module before the final segmentation.
\cite{ventura2019rvos} extends the spatial recurrence proposed for image instance segmentation \cite{salvador2017recurrent} with temporal recurrence, designing an architecture for zero-shot video object segmentation (without using the first ground-truth mask).

In this paper, we focus on using ConvLSTM for processing sequential information at multiple scales, following the ideas in \cite{xu2018youtub}.
In the next sections, we elaborate on our method and proceed with the implementation details followed by experiments as well as an ablation study on the impact of different components and the choice of hyper-parameters.
%
%
%
%
%
\section{Method}
In this section we describe our method including the modifications to the S2S architecture and the use of our multi-task loss for video object segmentation.
The S2S model is illustrated with yellow blocks in \autoref{fig:architecture} where the segmentation mask is computed as (adapted from \cite{xu2018youtub}):
\begin{equation}
    h_{0}, c_{0} = Initializer(x_{0}, y_{0})
\end{equation}
\begin{equation}
    \Tilde{x}_{t}= Encoder(x_{t})
\end{equation}
\begin{equation}
    h_{t}, c_{t} = RNN1(\Tilde{x_{t}}, h_{t-1}, c_{t-1})
\end{equation}
\begin{equation}
    \hat{y_{t}} = Decoder(h_{t})
\end{equation}
with $x$ referring to the RGB image and $y$ to the binary mask.
\subsection{Integrating Skip-Memory Connections}
We aim to better understand the role of the memory module used in the center of the encoder-decoder architecture in S2S method.
To this end, we replaced the ConvLSTM with simply feeding the previous mask as guidance for predicting the next mask, similar to \cite{perazzi2017learning}.
Doing so, we observed a drastic performance drop of about ten percent in the overall segmentation accuracy.
This suggests that only the guidance signal from the previous segmentation mask is not enough and that features from the previous time step should be aligned to the current time step.
As a result, we hypothesise that the role of ConvLSTM in the architecture is twofold: First, to remember the object of interest through the recurrent connections and the hidden state, and to mask out the rest of the scene and second, to align the features from the previous step to the current step, having a role similar to optical flow.
%

As mentioned earlier, the S2S model incorporates a memory module at the bottleneck of the encoder-decoder network. 
By inspecting the output of this approach, we noticed that the predicted masks for small objects are often worse than the other objects (see \autoref{fig:planes} and \autoref{fig:examples} for visual examples).
The issue is that the target object often gets lost early in the sequence as shown in \autoref{fig:uncertainty}.
We reason that this is partially due to the lack of information for smaller objects in the bottleneck.
For image segmentation, this issue is addressed via introducing skip connections between the encoder and the decoder \cite{ronneberger2015u, badrinarayanan2017segnet}.
This way the information about small objects and fine details are directly passed to the decoder.
Using skip connections is very effective in image segmentation;
however, when working with video, if the information in the bottleneck (input to the memory) is lost, the memory concludes that there is no object of interest in the scene anymore (since the memory provides information about the target object and its location).
As a result, the information in the simple skip connections will not be very helpful in this failure mode.

As a solution, we propose a system that keeps track of features at
different scales of the spatio-temporal data by using a ConvLSTM in the skip connection as shown in \autoref{fig:architecture}.
We note that some technical considerations should be taken into account when employing ConvLSTM at higher image resolutions.
As we move to higher resolutions (lower scales) in the video, the motion is larger and also the receptive field of the memory is smaller.
As stated in \cite{reda2018sdc}, capturing the displacement is limited to the kernel size in kernel-based methods such as using ConvLSTMs.
Therefore, adding ConvLSTMs at lower scales in the decoder, without paying attention to this aspect might have negative impact on the segmentation accuracy.
Moreover, during our experiments we observed that it is important to keep the simple skip connections (without ConvLSTM) intact in order to preserve the uninterrupted flow of the gradients.
Therefore, we add the ConvLSTM in an additional skip connection (RNN2 in \autoref{fig:architecture}) and merge the information from different branches using weighted averaging with learnable weights.
Hence, it is possible for the network to access information from different branches in an optimal way.

For the training objective of the segmentation branch in \autoref{fig:architecture}, we use the sum of the balanced binary-cross-entropy loss \cite{caelles2017one} over the sequence of length $T$, defined as:
\begin{equation}
  \begin{aligned} 
    L_{seg}(W) & = \sum_{t=1}^{T}(-\beta\sum\limits_{j\in Y_{+}}logP(y_{j}=1|X;\textbf{W})\\
      & -(1-\beta)\sum\limits_{j\in Y_{-}}logP(y_{j}=0|X;\textbf{W}))
  \end{aligned}
  \label{eq:seg_loss}
\end{equation}
where $X$ is the input, $\textbf{W}$ is the learned weights, $Y_{+}$ and $Y_{-}$ are foreground and background labeled pixels, $\beta=|Y_{-}|/|Y|$, and $Y$ is the total number of pixels.
\begin{figure}
    \centering
    \includegraphics[width=3in]{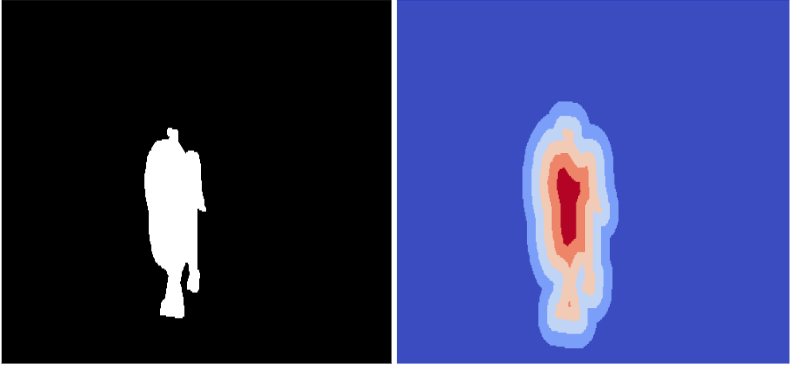}
    \vspace{1mm}
    \caption{In this figure, we show a binary mask (left) together with a heat-map depicting the distance classes (right).
    The number of distance classes is determined by two hyper-parameters for the \emph{number of border pixels} around the edges and the \emph{bin size} for each class. 
    The visualization shows that unlike previous works, our representations captures distance classes inside (reddish colors) as well as outside of the objects (blueish colors).
    }
    \label{fig:distance}
\end{figure}
\begin{table*}[!htbp]
\centering
\caption{Comparison of our best-obtained results with the state of the art approaches in video object segmentation using YoutubeVOS dataset. The values are reported in percentages and divided into columns for each score as in \cite{xu2018youtub}.
The table is divided to two parts for methods with and without online training.
We can see that our approach (even without online training) achieves the best overall score.
}
\begin{tabular}{lc c c c c c}
    Method & Online training & $J_{seen}$ & $J_{unseen}$ & $F_{seen}$ & $F_{unseen}$ & overall\\
\hline
OSVOS \cite{caelles2017one}& yes & 59.8 & \textbf{54.2} & 60.5 & \textbf{60.7} & \textbf{58.08}\\

MaskTrack \cite{perazzi2017learning} & yes & 59.9 & 45.0 & 59.5 & 47.9 & 53.08 \\

OnAVOS \cite{voigtlaender2017online}& yes & \textbf{60.1} & 46.6 & \textbf{62.7} & 51.4 & 55.20\\
\hline
OSMN \cite{yang2018efficient} & No & 60.0 & 40.6 & 60.1 & 44.0 & 51.18\\

RVOS \cite{ventura2019rvos} & No & 63.6 & 45.5 & 67.2 & 51.0 & 56.83\\

S2S \cite{xu2018youtub} & No & 66.7 & 48.2 & 65.5 & 50.3 & 57.68\\

S2S++(ours) & No & \textbf{68.68} & \textbf{48.89} & \textbf{72.03} & \textbf{54.42} & \textbf{61.00}\\
\hline
\end{tabular}
\label{tab:SOTA}
\end{table*}
\subsection{Border Distance Mask and Multi-Task Objective}
As the second extension, we build upon previous work of Bischke et al. \cite{bischke2019multi} and train the network parameters in addition to the object segmentation mask with an image representation based on a distance transformation (see \autoref{fig:distance} for an example).
This image representation was successfully used in a multi-task learning setup to explicitly bias the model to focus more on those pixels which are close to the object boundary and more error prone for misclassification, compared to the ones further away from the edge of the object. 

In order to derive this representation, we first apply the distance transform to the object segmentation mask. 
We truncate the distance at a given threshold to only incorporate the nearest pixels to the border. Let $Q$ denote the set of pixels on the object boundary and $C$ the set of pixels belonging to the object mask. For every pixel $p$ we compute the truncated distance $D(p)$ as: 
\begin{equation}
\begin{split} 
D(p) = \delta_p\  \inf \{\ \min_{\forall q \in Q}  d(p, q), R \ \}, \\
\mbox{where} \ \delta_p = \begin{cases}
+1 & \text{if} \quad p \in C \\
-1 & \text{if} \quad p \notin C
\end{cases}
\end{split}
\end{equation}
where $d(p,q)$ is the Euclidean distance between pixels $p$ and $q$ and $R$ is the maximal radius (truncation threshold).
The pixel distances are additionally weighted by the \emph{sign function} $\delta_p$ to represent whether pixels lie inside or outside objects. 
The continuous distance values are then uniformly quantized with a bin-size $s$ into $ \lfloor R/s \rfloor$ bins.
Considering both inside and outside border pixels, this yields to $2*R/s$ binned distance classes as well as two classes for pixel distances that exceeds the threshold R. 
We one-hot encode every pixel $p$ of this image representation into $k$ classification maps $D_K(p)$ corresponding each of the border distance.

%

We optimize the parameters of the network with a multi-task objective by combining the loss for the segmentation mask $L_{seg}$ and the loss for the border distance mask $L_{dist}$ as a weighted sum as follows.
Since we consider a multi-class classification problem for the distance prediction task we use the cross-entropy loss.
$L_{dist}$ is defined as the cross entropy loss between the derived distance output representation  $D_K(p)$ and the network output:
\begin{equation}
 L_{total} = \lambda\ L_{seg} + (1-\lambda)\ L_{dist}   
 \label{eq:total}
\end{equation}
The loss of the object segmentation task is the balanced binary-cross-entropy loss as defined in \autoref{eq:seg_loss}. The network can be trained end-to-end.
%
%
%
%
%
%
\section{Implementation Details}
In this section we describe implementation details of our approach.
\subsection{Initializer and Encoder Networks}
The backbone of the initializer and the encoder networks in \autoref{fig:architecture} is a VGG16 \cite{simonyan2014very} pre-trained on ImageNet \cite{krizhevsky2012imagenet}.
The last layer of VGG is removed and the fully-connected layers are adapted to a convolution layer to form a fully convolutional architecture as suggested in \cite{long2015fully}.
The number of input channels for the initializer network is changed to $4$, as it receives the RGB and the binary mask of the object as the input.
The initializer network has two additional $1\times1$ convolution layers with $512$ channels to generate the initial hidden and cell states of the ConvLSTM at the bottleneck (RNN1 in \autoref{fig:architecture}). 
For initializing the ConvLSTMs at higher scales, up-sampling followed by convolution layers are utilized, with the same fashion as the decoder. 
Additional convolution layers are initialized with Xavier initialization \cite{glorot2010understanding}.
\subsection{RNNs}
The ConvLSTM $1,2$ (shown as RNN1 and RNN2 in \autoref{fig:architecture}) both have a kernel size of $3\times3$ with $512$ channels. 
The ConvLSTM at the next level has a kernel size of $5\times5$ with $256$ channels.
Here, we chose a bigger kernel size to account for capturing larger displacements at lower scales in the image pyramid.

\subsection{Decoder} The decoder consists of five up-sampling layers with bi-linear interpolation, each followed by a convolution layer with kernel size of $5\times5$ and Xavier initialization \cite{glorot2010understanding}.
the number of channels for the convolution layers are $512, 256, 128, 64, 64$ respectively.
The features from the skip connections and the skip-memory are merged using a $1\times1$ convolution layer.
To adapt the decoder for the multi-task loss, an additional convolution layer is used to map $64$ channels to the number of distance classes.
This layer is followed by a $softmax$ to generate the distance class probabilities.
The distance scores are merged into the segmentation branch where a $sigmoid$ layer is used to generate the binary segmentation masks.

We use the Adam optimizer \cite{kingma2014adam} with an initial learning rate of $10^{-5}$.
In our experiments we set the value of $\lambda$ in Equation \ref{eq:total} to $0.8$.
When the training loss is stabilized, we decrease the learning rate by a factor of $0.99$ every $4$ epochs.
Due to GPU memory limitations, we train our model with batch size $4$ and a sequence length of $5$ to $12$ frames.
\begin{table*}[!htbp]
\centering
\caption{Ablation study on the impact of skip-memory and multi-task loss.
We can notice that multi-task loss and skip-memory individually improve the results, but lead to the best results when combined.
}
\begin{tabular}{c c c c c c}
    Method  & $J_{seen}$ & $J_{unseen}$ & $F_{seen}$ & $F_{unseen}$ & overall\\
\hline
base model & 65.36 & 43.55 & 67.90 & 47.50 & 56.08\\

base model + multi-task loss & 67.65 & 44.62 & 70.81 & 49.84 & 58.23\\
base model + one skip-memory & 66.89 & 46.82 & 69.22 & 50.08 & 58.25\\

base model + one skip-memory + multi-task loss & 67.18 & 47.04 & 70.24 & 52.30 & 59.19\\

base model + two skip-memory + multi-task loss & \textbf{68.68} & \textbf{48.89} & \textbf{72.03} & \textbf{54.42} & \textbf{61.00}\\

\hline
\end{tabular}
\label{tab:ablation}
\end{table*}
\begin{table*}[!htbp]
\centering
\caption{Results for different hyper-parameters for the multi-task loss on our best model. 
We can see that a higher number of distance classes slightly improves the metrics.}
\begin{tabular}{c c c c c c c c}
      border size & bin size & number of classes & $J_{seen}$ & $J_{unseen}$ & $F_{seen}$ & $F_{unseen}$ & overall\\
\hline
20 & 10 & 6 & 68.37 & 47.68 & 71.54 & 52.38 & 59.99 \\
20 & 1  & 42 & \textbf{68.68} & \textbf{48.89} & \textbf{72.03} & \textbf{54.42} & \textbf{61.00}\\
10 & 1  & 22 & 68.40 & 47.91 & 71.61 & 52.83 & 60.19\\
\hline
\end{tabular}
\label{tab:hyper}
\end{table*}
\begin{figure*}
    \centering
    \includegraphics[width=\textwidth]{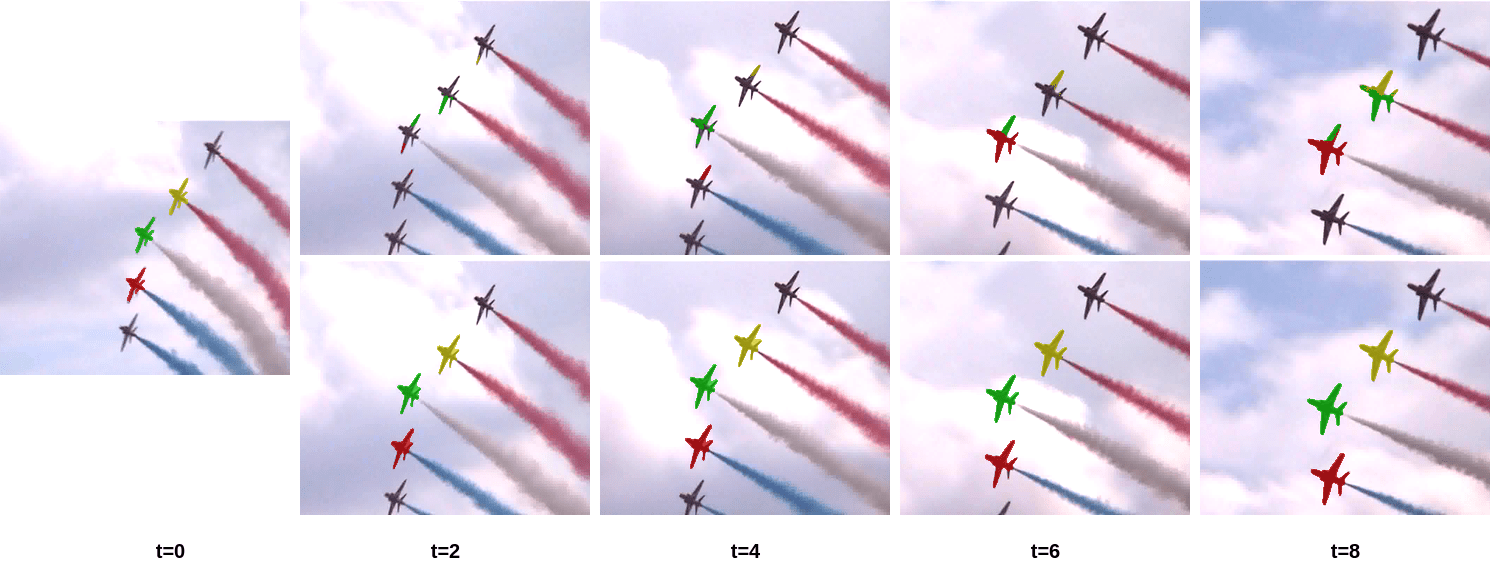}
    \caption{Qualitative comparison between the results obtained from S2S approach (first row) and the results from our method (second row).
    The first mask ($t=0$) is provided at test time and the target objects are segmented independently throughout the whole sequence.
    Every second frame is shown here and the brightness of the images is adjusted for better visibility.
    As it can be seen, our approach successfully tracks the target airplanes throughout the sequence while the S2S method loses and mixes the object masks early in the sequence.}
    \label{fig:planes}
\end{figure*}
\subsection{Border Output Representation}
The number of border pixels and the bin size per class are hyper-parameters which determine the resulting number of distance classes. 
In our internal experiments (see \autoref{subsec:ablation}), we noticed better results can be achieved if the number of distance classes is increased. 
In the following experiments, we set the \emph{border\_pixels}=20, the \emph{bin\_size}=1. Thereby we obtain for each object a segmentation mask with $42$ distance classes (the number of output classes is $2\times\frac{border\_pixels}{bin\_size}+2$).
Having the edge as the center, we have $\frac{border\_pixels}{bin\_size}$ classes at each of the inner and outer borders plus two additional classes for pixels which do not lie within the borders (inside and outside of the object) as shown in \autoref{fig:distance}.
\subsection{Data Pre- and Post-Processing} 
In line with the previous work in multiple object video segmentation, we follow a training pipeline, in which every object is tracked independently and at the end the binary masks from different objects are merged into a single mask.
For pixels with overlapping predictions, the label from the object with the highest probability is taken into account.
For data loading during the training phase, each batch consists of a single object from a different sequence.
We noticed that processing multiple objects of the same sequence degrades the performance.
The images and the masks are resized to $256\times448$ as suggested in \cite{xu2018youtub}.
For data augmentation we use random horizontal flipping and affine transformations.
For the results provided in \autoref{seq:results}, we have not used any refinement setp (e.g. CRF \cite{krahenbuhl2011efficient}) or inference-time augmentation. 
Moreover, we note that pre-training on image segmentation datasets can greatly improve the results due to the variety of present object categories in these datasets. However, in this work we have solely relied on pre-trained weights from ImageNet \cite{krizhevsky2012imagenet}.
%
%
%
%
%
\section{Experiments and Results}
\label{seq:results}
In this section a comparison with the state-of-the-art methods is provided in \autoref{tab:SOTA}. Additionally, we perform an ablation study in \autoref{tab:ablation} to examine the impact of skip-memory and multi-task loss in our approach. 

We evaluate our method on YoutubeVOS dataset \cite{xu2018youtub} which is currently the largest dataset for video object segmentation.
We use the standard evaluation metrics \cite{perazzi2016benchmark}, reporting \textit{Region Similarity} and \textit{Contour Accuracy} ($J \& F$).
$J$ corresponds to the average intersection over union between the predicted segmentation masks and the ground-truth, and $F$ is defined as $F = 2\frac{precision*recall}{precision+recall}$, regarding the boundary pixels after applying sufficient dilation to the object edges. 
For an overall comparability, we use the \textit{overall} metric of the dataset \cite{xu2018youtub} that refers to the average of $J \& F$ scores.
\subsection{Comparison to state-of-the-art approaches}
In \autoref{tab:SOTA}, we provide a comparison to state of art methods with and without online training.
As mentioned in \autoref{seq:related}, online training is the process of further training at test time through applying a lot of data augmentation on the first mask to generate more data.
This phase greatly improves the performance, at the expense of slowing down the inference phase.
As it can be seen in \autoref{tab:SOTA}, the scores are measured for two categories of seen and unseen objects.
This is a difference between other datasets and YoutubeVOS \cite{xu2018youtub} which consists of new object categories in the validation set.
Specifically, the validation set in YoutubeVOS dataset includes $474$ videos with 65 seen and 26 unseen categories. 
The score for unseen categories serves as a measure of generalization of different models.
As expected, the unseen object categories achieve a higher score when using online training (since the object is already seen by the network during the online training).
However, despite not using online training (and therefore also having lower computational demands during test time), S2S and S2S++ achieve higher overall performance. It is worth mentioning, that both $F_{seen}$ and $F_{unseen}$ scores improve by more than 4 percentage points in our approach.

\autoref{fig:planes} illustrates a qualitative comparison between our results and the ones from the S2S method. We provide additional examples in \autoref{fig:examples}.
\subsection{Ablation Study} \label{subsec:ablation}
Since the S2S method is the base of our work and the source code is not available, we provide a comparison between our implementation of S2S and S2S++ in \autoref{tab:ablation}, when adding each component. 
As it can be seen from the results, the best performance is achieved when using two skip-memory modules and multi-task loss.
We then take this model and experiment with different hyper-parameters for multi-task loss, as shown in \autoref{tab:hyper}.
The results show that a higher number of border classes that is closer to regression yields to a higher overall score.

\begin{figure*}
    \centering
    \includegraphics[width=\textwidth]{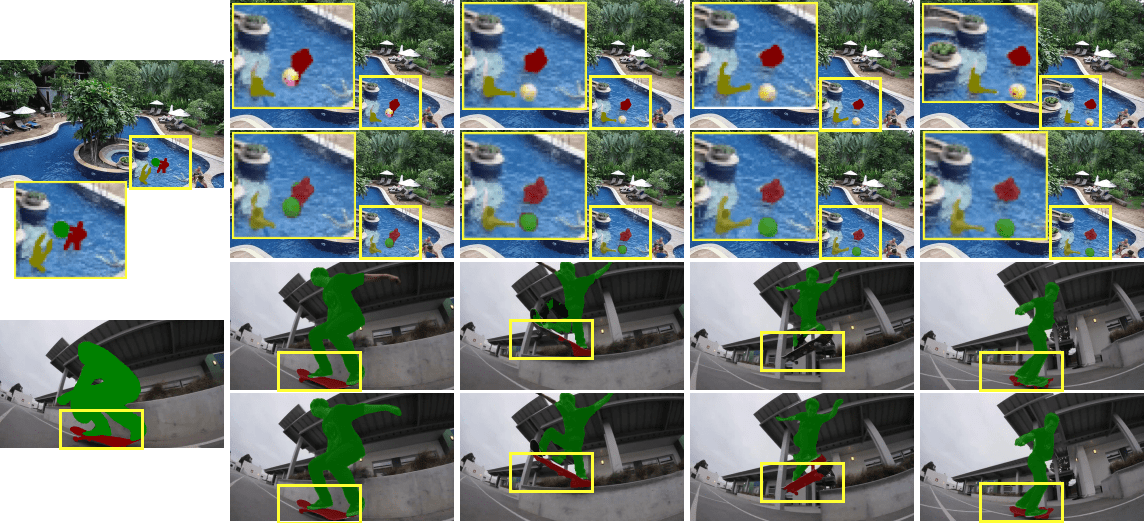}
    \caption{Additional examples for qualitative comparison between the S2S method in the first row, and ours in the second row.
    In the top part, the yellow frames were zoomed in for better visibility.
    We can observe a better capacity for tracking small objects.}
    \label{fig:examples}
\end{figure*}

It is worth mentioning that the distance loss ($L_{dist}$) has less impact for small objects, especially if the diameter of the object is below the border size (in this case no extra distance classes will be added).
Hence, we suspect the improvement in segmenting small objects (shown in \autoref{fig:planes} and \autoref{fig:examples}) is mainly due to the use of skip-memory connections.
%
%
%
%
%
%
\section{Conclusion}
In this work we observed that the S2S method often fails when segmenting small objects.
We build on top of this approach and propose using skip-memory connections for utilizing multi-scale spatio-temporal information of the video data.
Moreover, we incorporate a distance-based multi-task loss to improve the predicted object masks for video object segmentation.
In our experiments, we demonstrate that this approach outperforms state of the art methods on the \textit{YoutubeVOS} dataset \cite{xu2018youtub}. 
Our extensions to the S2S model require minimal changes to the architecture and greatly improves the contour accuracy score (\textit{F}) and the overall metric.

One of the limitations of the current model is a performance drop for longer sequences, especially in the presence of multiple objects in the scene.
In future, we would like to study this aspect and investigate the effectiveness of using attention as a potential remedy. 
Furthermore, we would like to study the multi task loss in more. One interesting direction is to learn separate task weights for the segmentation and distance prediction task as in \cite{bischke2019multi} rather than using fixed task weights as in our work. In this context, we would also like to examine the usage of a regression task rather than classification task for predicting the distance to the object border. 
\section*{Acknowledgments}
This work was supported by the TU Kaiserslautern CS PhD scholarship program, the BMBF project DeFuseNN (Grant 01IW17002) and the NVIDIA AI Lab (NVAIL) program.
Further, we thank all members of the Deep Learning Competence Center at the DFKI for their feedback and support. 

\bibliographystyle{plain}
\bibliography{my_bib}
    
\end{document}